\documentclass{ecai} 

\usepackage{latexsym}
\usepackage{amssymb}
\usepackage{amsmath}
\usepackage{amsthm}
\usepackage{booktabs}
\usepackage{enumitem}
\usepackage{graphicx}
\usepackage{color}
\usepackage{subcaption}
\usepackage{mathrsfs}

\newtheorem{theorem}{Theorem}

\newtheorem{example}[theorem]{Example}

\newcommand{\R}{\mathbb{R}}

\newcommand{\mb}[1]{\mathbf{#1}}

\newcommand{\bs}[1]{\boldsymbol{#1}}

\newcommand{\I}{\boldsymbol{I}}
\newcommand{\x}{\boldsymbol{x}}

\newcommand{\bl}{\begin{bmatrix}}
\newcommand{\br}{\end{bmatrix}}
\newcommand{\ml}{\begin{matrix}}
\newcommand{\mr}{\end{matrix}}

\newcommand{\sign}{\text{sign}}
\newcommand{\ff}[1]{\frac{1}{#1}}

\newcommand{\BibTeX}{B\kern-.05em{\sc i\kern-.025em b}\kern-.08em\TeX}

\begin{document}


\begin{frontmatter}


\paperid{2} 


\title{Multi-view mid fusion: a universal approach for learning in an HDLSS setting}


\author[A]{\fnms{Lynn}~\snm{Houthuys}\orcid{0000-0003-0867-4168}\thanks{Corresponding Author. Email: lynn.houthuys@vub.be.}}

\address[A]{Vrije Universiteit Brussel, Belgium}


\begin{abstract}
The high-dimensional low-sample-size (HDLSS) setting presents significant challenges in various applications where the feature dimension far exceeds the number of available samples. This paper introduces a universal approach for learning in HDLSS settings using multi-view mid fusion techniques. It shows how existing mid fusion multi-view methods perform well in an HDLSS setting even if no inherent views are provided. Three view construction methods are proposed that split the high-dimensional feature vectors into smaller subsets, each representing a different view. Extensive experimental validation across model-types and learning tasks confirm the effectiveness and generalization of the approach. We believe the work in this paper lays the foundation for further research into the universal benefits of multi-view mid fusion learning.
\end{abstract}

\end{frontmatter}


\section{Introduction}

The \emph{high-dimensional low-sample-size} (HDLSS) setting concerns applications where the feature dimension is (much) higher than the number of samples available. This is a common issue in applications such as radiomics \cite{Cao2019}, microarray data analysis \cite{Vatankhah2024}, astronomy and atmospheric science \cite{Johnstone2009} and even computer vision \cite{Chen2000}.
The HDLSS problem poses a challenge to various machine learning models. The most common issues are related to the curse of dimensionality due to the collapse of pairwise distances in high dimensions \cite{Cai2024}, overfitting \cite{Shen2022, Qiao2015}, noise accumulation \cite{Fan2014a} and overall poor classification accuracy (sometimes referred to as the Hughes problem \cite{Salimi2018}). Specifically for Neural Network (NN)-type models, the HDLSS setting causes high variance gradients \cite{Liu2017}, where calculating them can be very time consuming \cite{Fan2014a}. While kernel-based models have shown to generalize better in the HDLSS setting than their NN counterparts \cite{Salimi2018}, they tend to suffer from data-piling \cite{Marron2007}, especially when the classification problem is imbalanced \cite{Qiao2015}.

The most common approach to deal with HDLSS data is to first perform dimensionality reduction. However, if most features are relevant this approach will inevitably lose information, and if the sample size is very small, only very few features will be retained \cite{Shen2022}. Furthermore, finding the most optimal subset of features represents a combinatorial optimization problems that is NP-hard \cite{Charikar2000}. Finally, these methods are often time-consuming and impractical \cite{Shen2020} and can be seen as more of a workaround than a true solution \cite{Cavalheiro2024}.

\emph{Multi-view learning} describes a learning paradigm where data is represented by multiple \emph{views}. Most multi-view models can be roughly divided into three categories based on when information from different views is combined: \emph{early, late and mid fusion} (see Figure \ref{fig:fusions}). Various multi-view variants of traditional machine learning models have been proposed, and they usually show to generalize better than their single-view counterparts by taking into account information from all views. For a thorough overview of multi-view learning we refer to the work of Sun et al. \cite{Sun2019book}.

Most multi-view research assumes that the data is inherently described through multiple views. Sun \cite{Sun2013} stated that even when an inherent multi-view structure does no exist, performance improvements can still be observed using manufactured views. Views can be constructed by increasing the feature space, such as adding multiple representations (e.g. the work of Michaeli et al. \cite{Michaeli2016}) or by \emph{feature set partitioning} \cite{Sun2019book}, which splits a high dimensional feature vector into smaller sets which represent different views. 

While it is often assumed (such as by Di \& Crawford \cite{Di2012}) that splitting a high dimensional input into multiple views can alleviate issues like the curse of dimensionality, to the best of our knowledge it has never been studied thoroughly.
Some research has shown improvement for a specific model and/or application (e.g. by Sun et al. \cite{Sun2011}), but these results are not able to generalize to a more universal setting. Moreover most of the related research focusses on late fusion strategies and do not specifically take into account the HDLSS setting.

The aim of this work is to show that multi-view learning, and in particular mid fusion approaches, can overcome at least some of the challenges of the HDLSS problem. This paper aims to transcend model or learning problem-specific results, and shows the universality of the idea. 
The main contributions of this paper are:
\begin{itemize}
	\item Introducing a universal approach for learning in an HDLSS setting using multi-view mid fusion techniques
	\item Comparison and evaluation of early, mid and late fusion strategies in an HDLSS setting. It further highlights the advantage of mid fusion in terms of performance.
	\item Exploration of three different feature set partitioning methods and providing insights in how to use these for view construction to enhance performance in an HDLSS setting
	\item An extensive experimental study spanning multiple types of models and two types of learning problems. This includes kernel- and NN-based models and both classification and clustering. The results demonstrate the effectiveness of the proposed approach across different scenarios.

\end{itemize}


\subsection{Related work}

As mentioned above, a common way to deal with an HDLSS problem is by reducing the dimension by feature selection or feature transformation methods as a pre-processing step. Liu et al. \cite{Liu2017} was one of the first to apply deep learning on HDLSS data by using a feature selection procedure. Cao et al. \cite{Cao2019} consider a multi-view HDLSS setting, specifically in radiomics, and fuse representations of each view to a reduced dimensional vector representation. More recently, Mandal et al. \cite{Mandal2024} used an ensemble of feature selection methods to find the most optimal feature subset for HDLSS classification.

Most methods that have been proposed for HDLSS data without dimensionality reduction are kernel-based models based on Support Vector Machines (SVM) \cite{Vapnik1995}.  Such as the distance-weighted SVM \cite{Qiao2015}, bias-corrected nonlinear SVM \cite{Nakayama2020} and more recently the HDLSS-compliant similarity measure kernel for SVM \cite{Cavalheiro2024}. Most of these, however, still suffer from overfitting (usually illustrated by the data-piling phenomenon \cite{Marron2007}) and can under perform in certain conditions such as when there is class imbalance \cite{Qiao2015} or when the underlying distributions are close in their location and scale \cite{Roy2022}. Almost all of these are limited to binary classification. 

In recent years a lot of research has been done in how to efficiently deal with hyperspectral data \cite{Jia2021}. While this type of data is inherently high-dimensional, most of proposed methodologies do not consider the more challenging HDLSS setting. Multi-view learning via view construction is sometimes considered in hyperspectral data modelling, however, it is almost always handled by an ensemble approach (e.g. in \cite{Jamshidpour2020} and \cite{Di2012}). A recent review paper by Li et al. \cite{Li2022} provides an overview on multi-view learning for hyperspectral image classification, together with several view construction methods.


The most closely related paper is probably the work of Roy et al. \cite{Roy2022}, which introduced a grouped distance metric for binary classification in an HDLSS setting by dividing the features in groups that are alike. While this enforces the conceptual hypothesis of this paper that splitting up the feature dimension into views can improve learning, it differs in several key areas. More specifically: (1) it focusses on binary classification using Euclidean distance based classifiers, while we consider a variety of methods and learning problems; (2) it defines a new model, while we show that the existing mid fusion models already perform very well; (3) it uses an agglomerative hierarchical clustering method for feature set partitioning while we consider three more simple and light-weighted partitioning methods; (4) it introduces a grouped distance metric while this work uses mid fusion multi-view methods which can capture more complex view relations.


\section{Methodology}

Consider a training set of $N$ data points $\{\x_k, y_k\}_{k=1}^N$ where $\x_k \in \R^d$ is a $d$-dimensional vector representation of the $k$-th input sample $x_k \in \mathcal{X}$ from the input space $\mathcal{X}$ . $y_k \in \mathcal{Y}$ denotes the label corresponding to the $k$-th sample, which can represent values from a discrete set (e.g. $\mathcal{Y} = \{-1,1\}$) for classification problems, from a continuous range (e.g. $\mathcal{Y} = \R$) for regression, or may not be (fully) available in the case of unsupervised learning problems.
We consider the learning problem in an HDLSS setting when $N<d$. Often we will even consider the case where the number of data points is drastically smaller than the dimension, i.e. $N\ll d$.

In order to improve learning in HDLSS settings, we propose to split the large feature vectors into smaller subsets called \emph{views}, and use mid fusion multi-view learning methods. Instead of using one large feature vector $\x_k \in \R^d$ for each sample $x_k$, we will use a multi-view representation of the data where each sample $x_k$ is represented by $V$ feature vectors $\x^{[v]}_k \in \R^{d_v}$ for $v=1,\ldots,V$ where $d_1 + \ldots + d_V=d$.

Often HDLSS data is already inherently multi-view. E.g. multi-omics data \cite{Kang2022, Wang2021} is a collection of various type of features such as mRNA expressions and DNA methylation, and each of these can be considered a different view. Even when the views are not specified, or even when a natural feature split does not exist, we can construct multiple views by feature set partitioning.

\subsection{Feature set partitioning}
\label{sec:ViewCon}

When no inherent views are available, we can still split up the features into disjoint subset, each of which represents one view. Suppose that $\mathscr{I}=\{1,2,\ldots,d\}$ represents the index set of features corresponding to each each data point.
The view construction problem can be formally defined as partitioning the index set into $V$ disjoint subsets, i.e., $\mathscr{I}=\cup_v \mathscr{I}_v$ and $\mathscr{I}_v \cap \mathscr{I}_u = \emptyset$, for $v\neq u$.

In this paper we considered three feature set partitioning methods: random split, Euclidean distance-based feature clustering and Correlation-based feature clustering. Note that all three methods are unsupervised, i.e. the labels are not taken into consideration for the construction of views. This makes them suitable for unsupervised and semi-supervised learning as well. Moreover, all are fairly light-weight in terms of complexity, making them suitable for real life scenarios.

\subsubsection{Random split}

The easiest way to construct views is to randomly partition the feature set. Although simple, this has shown promising results in the past for co-training style multi-view methods \cite{Brefeld2005, Nigam2000} outside of the HDLSS setting. An added advantage is that this does not significantly increases the training or inference time.

\subsubsection{Euclidean distance-based feature clustering}

We know from previous research (e.g. \cite{HOUTHUYS2018b}) that multi-view learning works best when the views are diverse enough, i.e. when each view contains specific information that is not present in the other views. We can leverage this insight by clustering the features into sets that are similar, such that the dissimilarity between the views is maximized. 

In this paper we used $k$-means \cite{Macqueen1967} with the Euclidean distance as a dissimilarity measure to cluster the features based on the training data (without the labels). This method was chosen because it is fast, easy to implement and one of the most well known clustering algorithms. Furthermore $k$-means was also used by Di \& Crawford \cite{Di2012} for view generation for hyperspectral data.

\subsubsection{Correlation-based feature clustering}

Inspired by the collinear group detection algorithm designed by Kim \& Kim \cite{Kim2018}, the third view construction method aims to find groups of features which are highly (linearly) correlated with each other. Instead of using the variance inflation factor, which is not possible when $d>N$, we use a clustering technique to approximate the optimal solution. 
More concretely, we use $k$-means to cluster the correlation matrix of the features. This way, the algorithm finds sets of features which are correlated similarly to all other features. This approach was successfully used before for feature extraction for hyperspectral data by Volpi et al. \cite{Volpi2014}.

Another approach could have been to use hierarchical clustering that takes the pairwise correlations as a similarity measure (similarly to the methodology by Roy et al. \cite{Roy2022}). While this alternative clusters the features more explicitly based on the correlations between them, it is much more computationally expensive. 

\subsection{Multi-view fusion}

\begin{figure*}
    \centering
    \begin{subfigure}[b]{0.3\textwidth}
        \includegraphics[width=\textwidth]{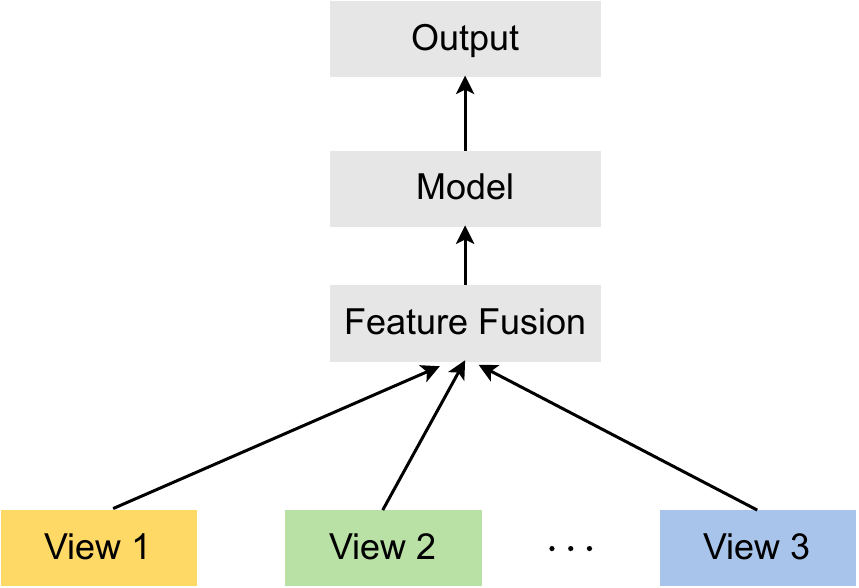}
        \caption{Early fusion}
        \label{fig:EarlyFusion}
    \end{subfigure}
    ~ 
    \begin{subfigure}[b]{0.3\textwidth}
        \includegraphics[width=\textwidth]{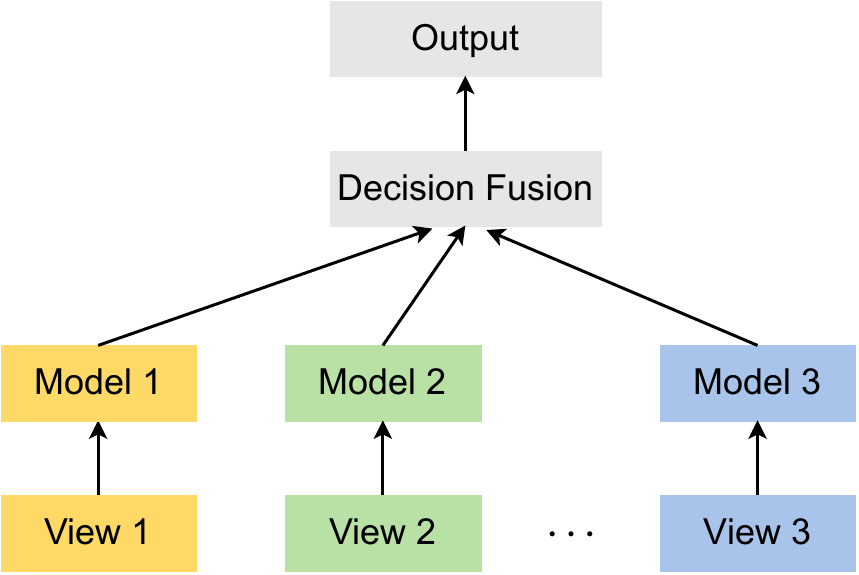}
        \caption{Late fusion}
        \label{fig:LateFusion}
    \end{subfigure}
    ~ 
    \begin{subfigure}[b]{0.3\textwidth}
        \includegraphics[width=\textwidth]{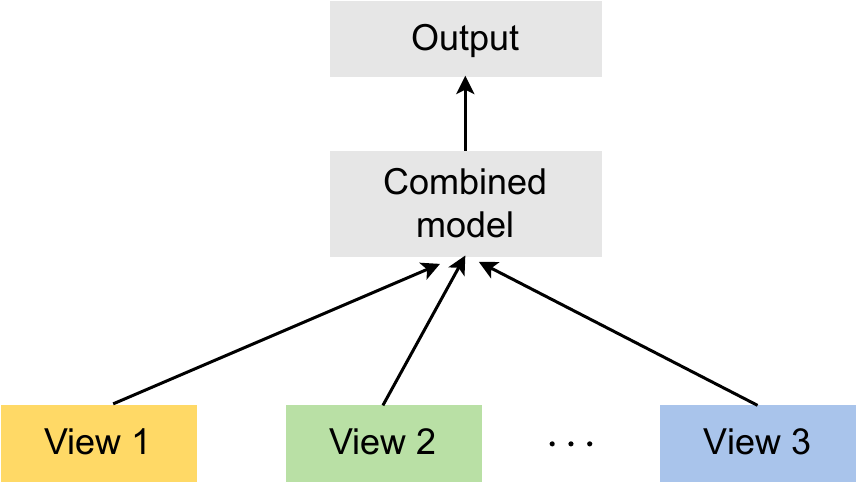}
        \caption{Mid fusion}
        \label{fig:MidFusion}
    \end{subfigure}
		
		\vspace{0.3cm}		
		
    \caption{Different types of multi-view fusion}\label{fig:fusions}
		\vspace{0.4cm}
\end{figure*}

The information from different views can be combined at different stages of the learning process, as visualized by Figure \ref{fig:fusions}.

\subsubsection{Early fusion}

In \emph{early fusion} (see Figure \ref{fig:EarlyFusion}), or feature-level fusion, the information from multiple views is combined before any training process is performed. A typical example is feature concatenation, where the features from each view are simply concatenated to form one large feature vector. 
Early fusion was used e.g. by Poria et al. \cite{Poria2015} for multi-modal sentiment analysis and Karevan et al. \cite{Karevan2015} to perform temperature prediction based on measurements in multiple cities. 
While this ensures that the information from all views is taken into account early on, it has some drawbacks. It ignores the statistical properties of each individual view and may not fully exploit the complementary nature of the views.

In our work, the original HDLSS problem where the learning model uses the large feature vector $\x_k \in \R^d$ directly, corresponds to this early fusion setting.

\subsubsection{Late fusion}

In \emph{late fusion} (see Figure \ref{fig:LateFusion}), or decision-level fusion, view-specific models are separately trained and the final decision is a combination of the view-specific decisions. Various rules exist to determine how decisions from different views are combined. While this technique ensures a high degree of freedom to model the views differently, the information from other views is not taken into account during the actual view-specific training. Most ensemble learners fall into this category, such as the multi-view method from Jamshidpour \cite{Jamshidpour2020} and Kumar \& Minz \cite{Kumar2016} for hyperspectral and HDLSS data respectively.

In our work we will use ensembles of well known methods, each trained independently on a separate view. 
Assume $V$ models $f_v: \R^{d_v} \rightarrow \mathcal{Y}$, each trained on a different view $v$ for $v=1,\ldots,V$. The final decision $\hat{y}_k(\x_{\mb{t}})$ (e.g. class label, cluster assignment etc.) for input $\x_{\mb{t}}=[\x_{\mb{t}}^{[1]} \ldots \x_{\mb{t}}^{[V]}]$ is made based on the view-specific outcomes using a decision rule $d: \mathcal{Y}^V \rightarrow \mathcal{Y}$ where 
\begin{equation}
\hat{y}_k(\x_{\mb{t}}) = d(f_1(\x_{\mb{t}}^{[1]}),\ldots,f_V(\x_{\mb{t}}^{[V]}))
\label{eq:lf}
\end{equation}
We consider two decision rules: average fusion and performance-based fusion. 

In \emph{average fusion}, the final decision is based on the mean prediction of the separate views. 
For classification and clustering this can be done at the level of the class labels or cluster assignment itself, which corresponds to what is sometimes called \emph{majority voting}. In this case $d$ equals taking the mode. When we have access to predicted probabilities (such as for NN-type methods with a sigmoid or softmax output layer) it makes sense to perform the fusion at that level, i.e. before the threshold is applied. 

In \emph{performance-based fusion}, the final decision is a weighted average of the view-specific predictions, where the weight is proportional to its performance on the training set or a separate validation set. E.g. for binary classification using an NN architecture with a sigmoid output layer: 
\begin{equation}
\hat{y}_k(\x_{\mb{t}}) = f(\beta_1 f_1(\x_{\mb{t}}^{[1]})+ \ldots + \beta_V f_V(\x_{\mb{t}}^{[V]}))
\label{eq:lf_NN}
\end{equation}
where $f$ is a threshold function that maps the weighted average of the predictions to class 0 or 1. In the same way as e.g. for Committee LS-SVM regression described by Suykens et al. \cite{Suykens2002}, the weights are chosen such that it minimizes the entire ensemble training error. This is achieved via the training error covariance matrix $\bs{C} \in \R^{V \times V}$ with $C_{vu}=\ff{N} \sum_{k=1}^N(f_v(\x_k^{[v]})-y_k)(f_u(\x_k^{[u]})-y_k)$, 
where 
\begin{equation}
\beta_v=\frac{\sum_u^V C_{vu}^{-1}}{\mb{1}_V^T \bs{C}^{-1} \mb{1}_V} \quad \text{for} \quad v=1,\ldots,V
\label{eq:}
\end{equation}
 and $\mb{1}_V \in \R^{V\times 1}$ is a vector of ones.

\subsubsection{Mid fusion}

\emph{Mid fusion} (see Figure \ref{fig:MidFusion}), or intermediate fusion, offers more flexibility as to how and when information from different views is fused. Typical mid fusion strategies are co-training \cite{Blum1998} and co-regularization. The latter adds up the different objective functions and introduces an extra regularization
term. In our work we will use two co-regularization style multi-view mid fusion methods, MV-LSSVM \cite{HOUTHUYS2018b} for kernel-based classification and Co-regularized Multi-view Spectral Clustering \cite{Kumar2011} for clustering. 
From an NN perspective, mid fusion uses view-specific layers, shared layers and a shared representation layer \cite{Ramachandram2017}.


\section{Experiments}

We performed three sets of experiments, each with a different type of model and/or learning task: (1) kernel-based classification, (2) NN-based classification and (3) spectral clustering.

For each set we picked a method that has a multi-view mid fusion variant. We compared the performance of the single-view method on all features concatenated (early fusion), the multi-view method (mid fusion) and using the single-view method on all views separately followed by using a joint decision function (late-fusion). We evaluated the methods for all three types of view construction methods described in Section \ref{sec:ViewCon}, while the number of views ($V\in\{2,3,4,5\}$) and the nature of the joint decision function (average or performance-based fusion) were part of the tuning process. When reporting the results we will use the abbreviations `EF', `MF' and `LF' for respectively early, mid and late fusion, and `inherent', `random', `kmeans' and `corr' for the feature partitioning methods used for view construction: inherent view structure, random split, Euclidean distance-based feature clustering and Correlation-based feature clustering, respectively. 

For each set of experiments we used the following examples to generate synthetic datasets, where the first two are designed for classification and the last two for clustering. Example \ref{ex_2} and Example \ref{ex_4} have an inherent multi-view structure.

\begin{example}
\label{ex_1}
The first example consist of simulated data for a binary classification problem, typically used in HDLSS research (e.g. by Shen et al. \cite{Shen2022} and Qiao \& Zhang \cite{Qiao2015}). The two classes are sampled from multivariate normal distributions $N_d(\pm \bs{\mu}, \bs{\Sigma})$, where $\bs{\mu}=a \bs{1_d}$, $\bs{\Sigma}=\I_{\bs{d}}$ and $a$ is a scaling factor with $2a||\bs{1_d}||_2=2.7$. This corresponds to  the Mahalanobis distance between the two classes and represents a reasonable difficulty of classification. 
\end{example}

\begin{example}
\label{ex_2}
The second example consist of simulated data for a binary classification problem, with an inherent multi-view structure. This example is similar to the examples used by Roy et al. \cite{Roy2022}. The two classes are sampled from multivariate normal distributions $N_d(\bs{\mu_1}, \bs{\Sigma_1})$ and $N_d(\bs{\mu_2}, \bs{\Sigma_2})$, where $\bs{\mu_1}=\bs{\mu_2}=\bs{0_d}$ and $\bs{\Sigma_1}$ and $\bs{\Sigma_2}$ are block diagonal matrices of the form:
$$\bs{\Sigma_c}=\bl \bs{\Sigma_c'} & 0 & \cdots & 0 \\ 0 & \bs{\Sigma_c'} & \cdots & 0 \\ \vdots & \vdots & \ddots & \vdots \\ 0 & \cdots & 0 & \bs{\Sigma_c'} \br \text{ with } \bs{\Sigma_c'}=\bl 1 & \tau_c & \cdots & \tau_c \\ \tau_c & 1 & \cdots & \tau_c \\ \vdots & \vdots & \ddots & \vdots \\ \tau_c & \tau_c & \cdots & 1 \br$$
for $c=1,2$ and where $\tau_1=0.3$, $\tau_2=0.7$ and $\bs{\Sigma_c'} \in \R^{d/5 \times d/5}$. Hence there are inherently 5 groups of features (=views).
\end{example}

\begin{example}
\label{ex_3}
The first clustering example uses data similar to Example \ref{ex_1}, but without class labels. Furthermore, in order to get meaningful results we increased the distance between the means of the two distributions by a factor of 2. Hence this corresponds to twice the Mahalanobis distance between the two clusters. 
\end{example}

\begin{example}
\label{ex_4}
The second clustering example uses data similar to Example \ref{ex_2}, but without class labels. Similarly to the previous clustering example, we increased the distance between the two clusters by taking $\bs{\mu_1}=-0.5$ and $\bs{\mu_2}=0.5$.
\end{example}

The number of training samples in all examples equal $N_1=120$ and $N_2=90$, and for the classification examples we consider a large test set of 3000 samples (1500 for each class). The dimension $d$ is varied from the set $\{80, 150, 240, 650, 900, 1500, 2400\}$, hence the last five cases correspond to the HDLSS definition.

For each set of experiments we sampled the datasets $10$ times, and tuned, trained and tested the different models. We used Random Search with $10$-fold cross-validation for the tuning process. The code from the experiments will be made available via GitHub after acceptance.

\begin{figure*}
    \centering
    \begin{subfigure}[b]{0.3\textwidth}
        \includegraphics[width=\textwidth]{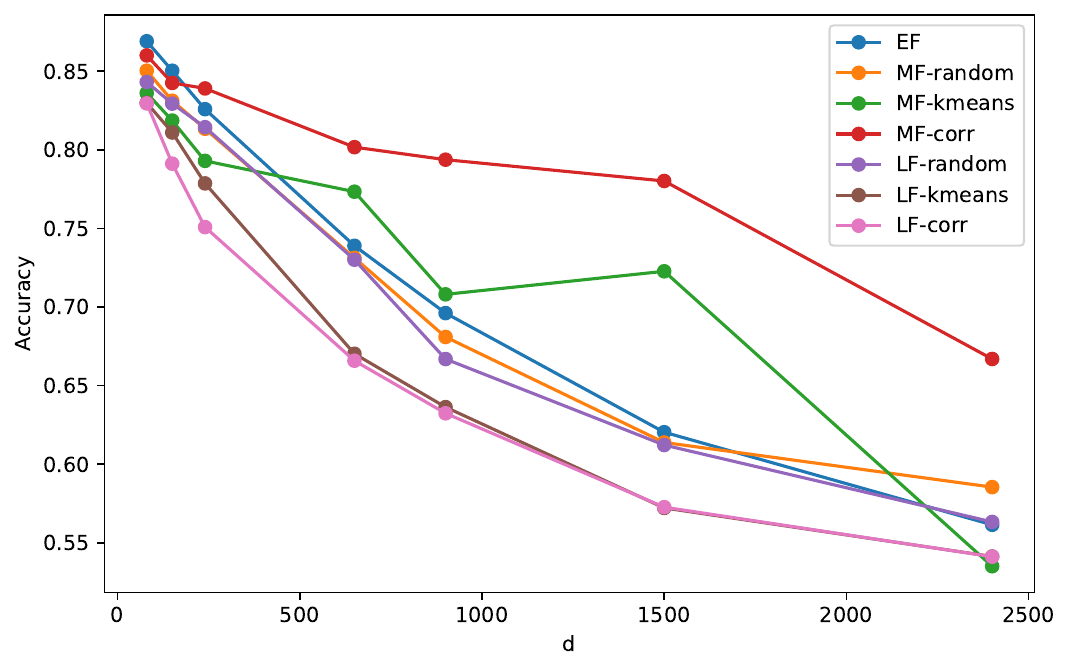}
        \caption{Mean accuracy of the kernel-based models on the test sets of Example \ref{ex_1}.}
				\label{fig:kernel_acc}
    \end{subfigure}
    ~ 
    \begin{subfigure}[b]{0.3\textwidth}
        \includegraphics[width=\textwidth]{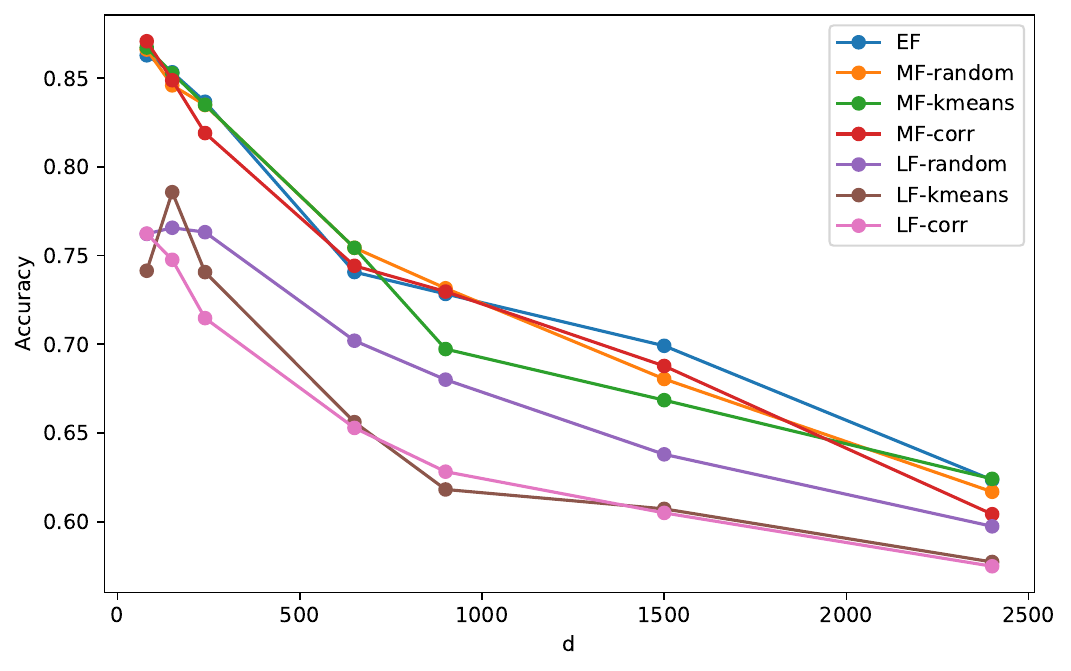}
        \caption{Mean accuracy of the NN-based models on the test sets of Example \ref{ex_1}.}
				\label{fig:NN_acc}
    \end{subfigure}
    ~ 
    \begin{subfigure}[b]{0.3\textwidth}
        \includegraphics[width=\textwidth]{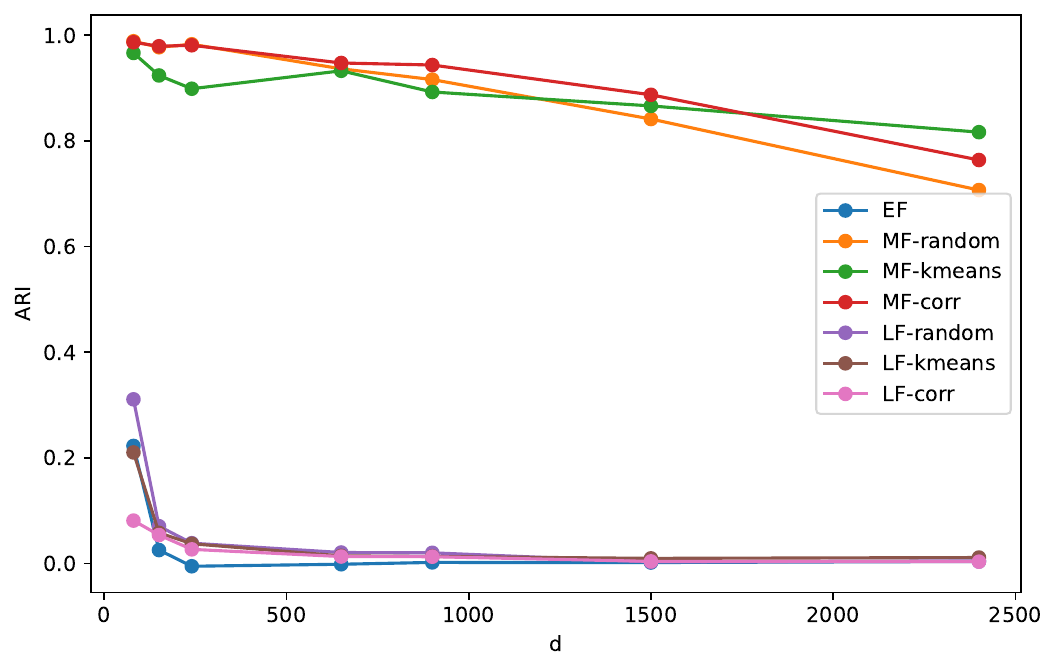}
        \caption{Mean ARI of the clustering models on the data of Example \ref{ex_3}.}
				\label{fig:SC_acc}
    \end{subfigure}
		
		\vspace{0.5cm}		
		
		\begin{subfigure}[b]{0.3\textwidth}
        \includegraphics[width=\textwidth]{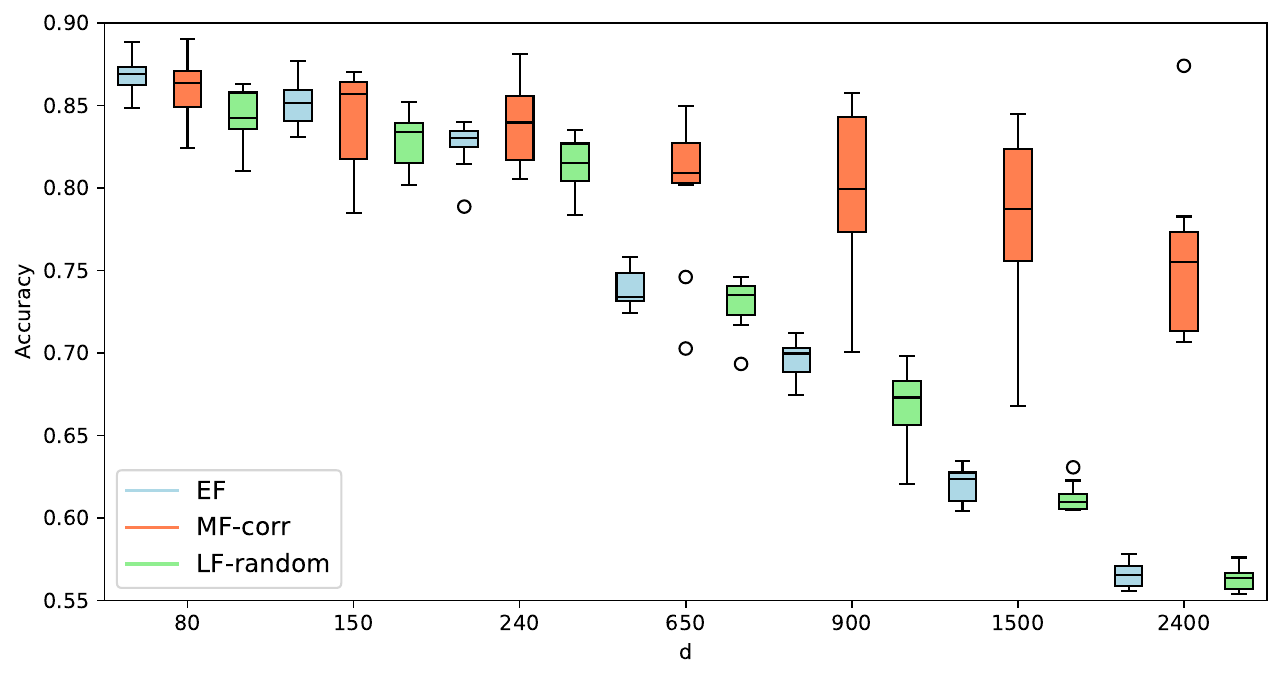}
        \caption{Accuracy of the kernel-based models on the test sets of Example \ref{ex_1}.}
				\label{fig:kernel_acc_boxp}
    \end{subfigure}
		~ 
    \begin{subfigure}[b]{0.3\textwidth}
        \includegraphics[width=\textwidth]{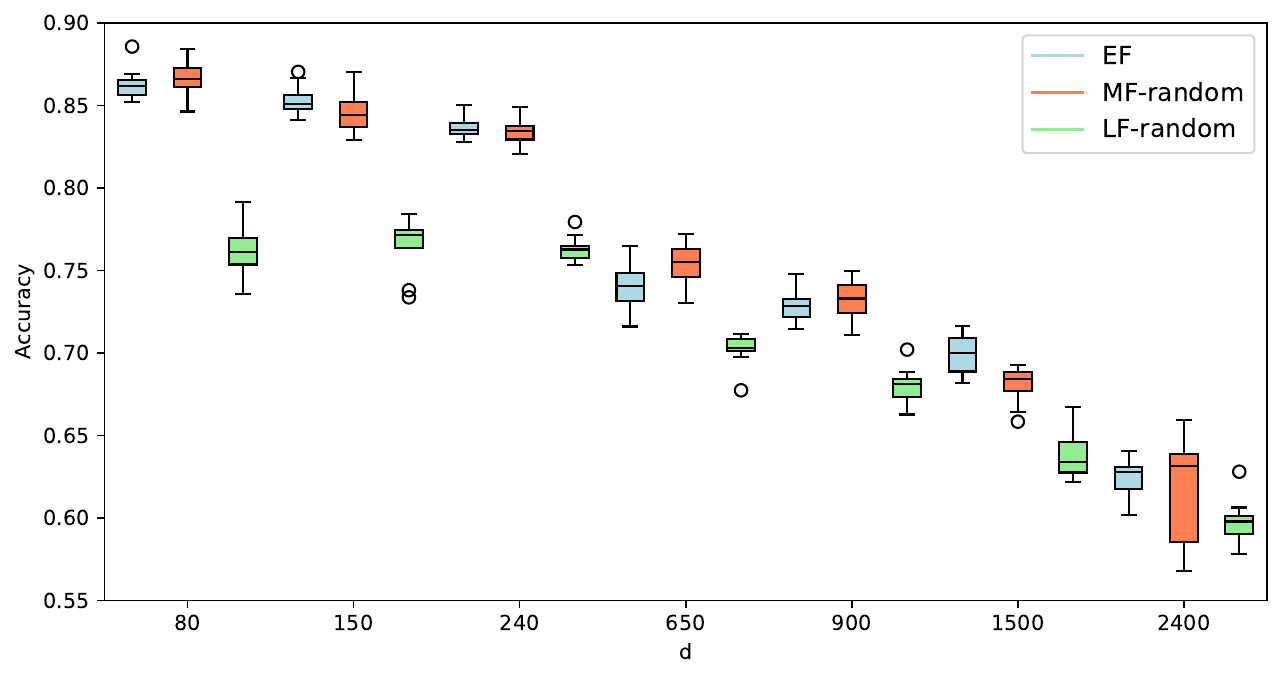}
        \caption{Accuracy of the NN-based models on the test sets of Example \ref{ex_1}.}
				\label{fig:NN_acc_boxp}
    \end{subfigure}
    ~ 
		\begin{subfigure}[b]{0.3\textwidth}
        \includegraphics[width=\textwidth]{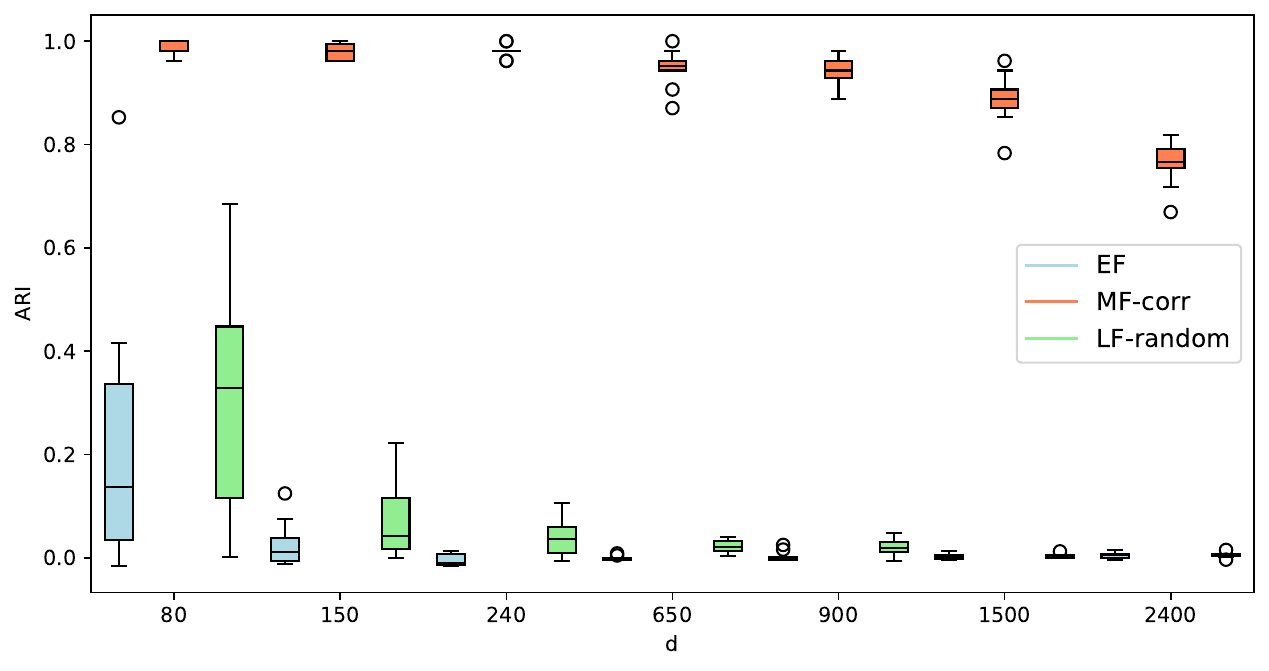}
        \caption{ARI of the clustering models on the data of Example \ref{ex_3}.}
				\label{fig:SC_boxp}
    \end{subfigure}
		
		\vspace{0.5cm}		
		
		\begin{subfigure}[b]{0.3\textwidth}
        \includegraphics[width=\textwidth]{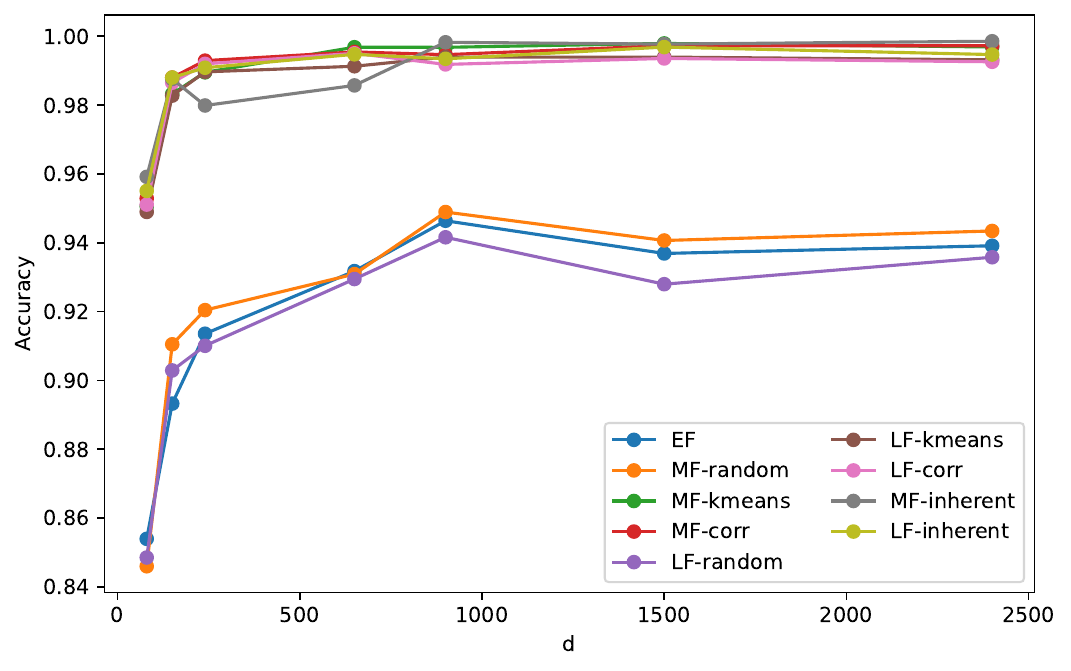}
        \caption{Mean accuracy of the kernel-based models on the test sets of Example \ref{ex_2}.}
				\label{fig:kernel2_acc}
    \end{subfigure}
    ~ 
    \begin{subfigure}[b]{0.3\textwidth}
        \includegraphics[width=\textwidth]{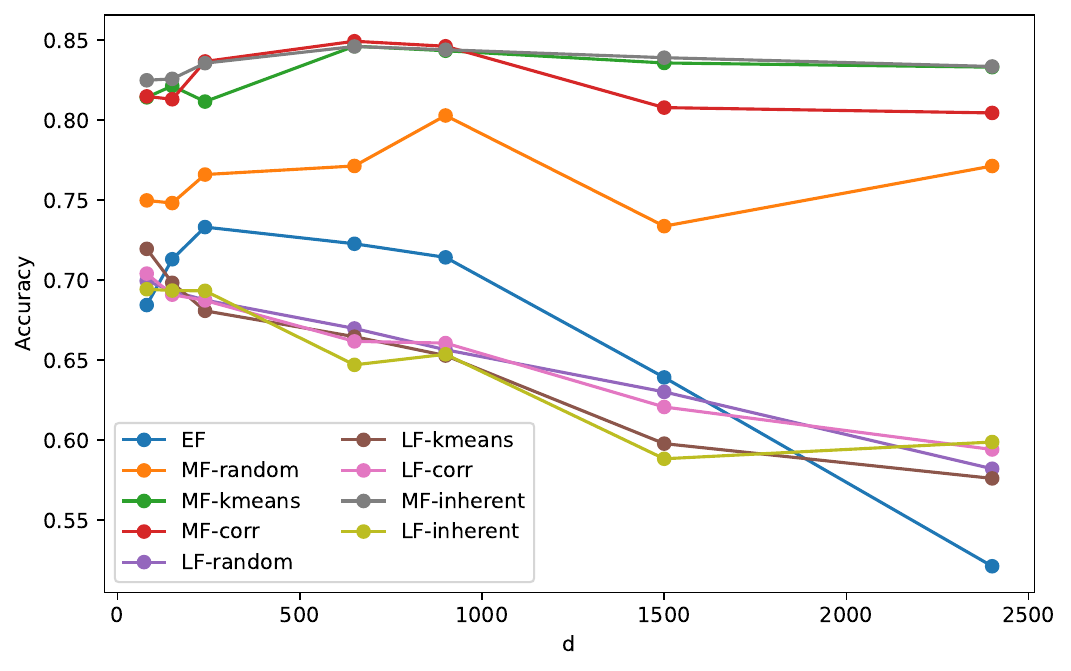}
        \caption{Mean accuracy of the NN-based models on the test sets of Example \ref{ex_2}.}
				\label{fig:NN2_acc}
    \end{subfigure}
    ~ 
    \begin{subfigure}[b]{0.3\textwidth}
        \includegraphics[width=\textwidth]{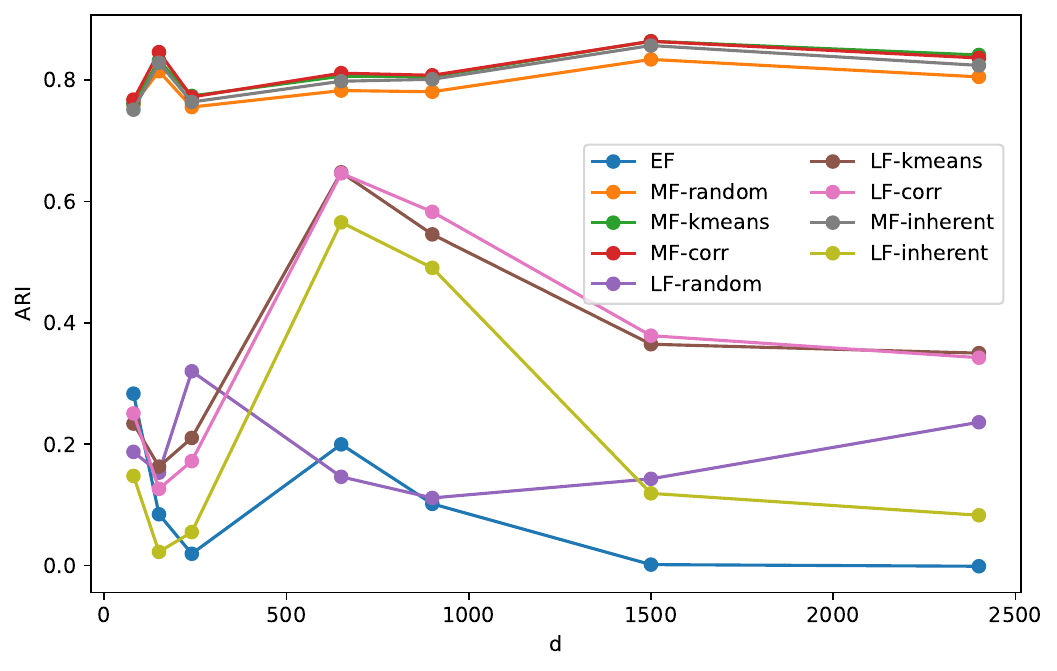}
        \caption{Mean ARI of the clustering models on the data of Example \ref{ex_4}.}
				\label{fig:SC2_acc}
    \end{subfigure}
		
		\vspace{0.5cm}		
		
		\begin{subfigure}[b]{0.3\textwidth}
        \includegraphics[width=\textwidth]{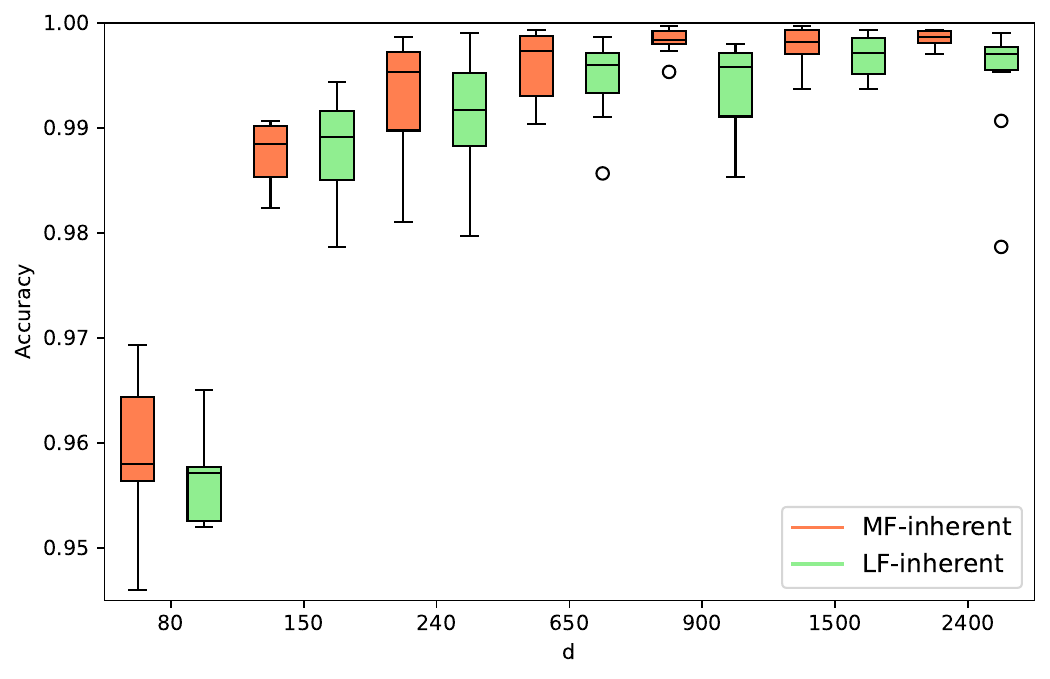}
        \caption{Accuracy of the kernel-based models on the test sets of Example \ref{ex_2}.}
				\label{fig:kernel2_acc_boxp}
    \end{subfigure}
		~ 
    \begin{subfigure}[b]{0.3\textwidth}
        \includegraphics[width=\textwidth]{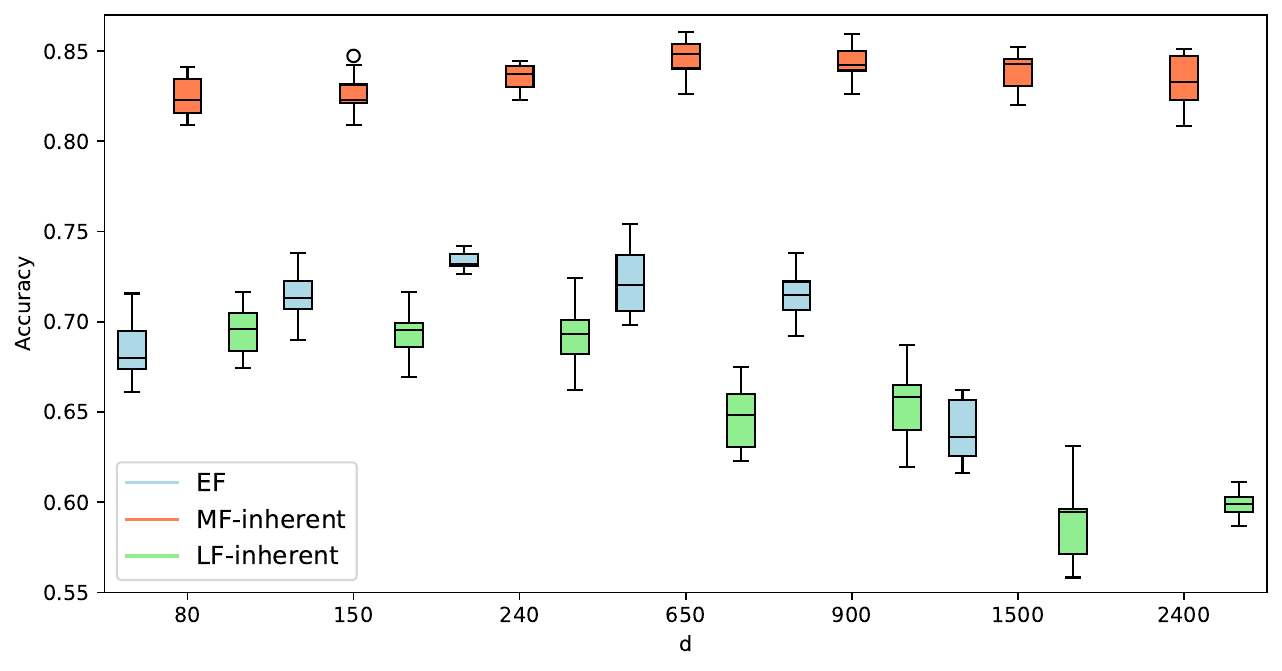}
        \caption{Accuracy of the NN-based models on the test sets of Example \ref{ex_2}.}
				\label{fig:NN2_acc_boxp}
    \end{subfigure}
    ~ 
    \begin{subfigure}[b]{0.3\textwidth}
		\centering
        \includegraphics[width=\textwidth]{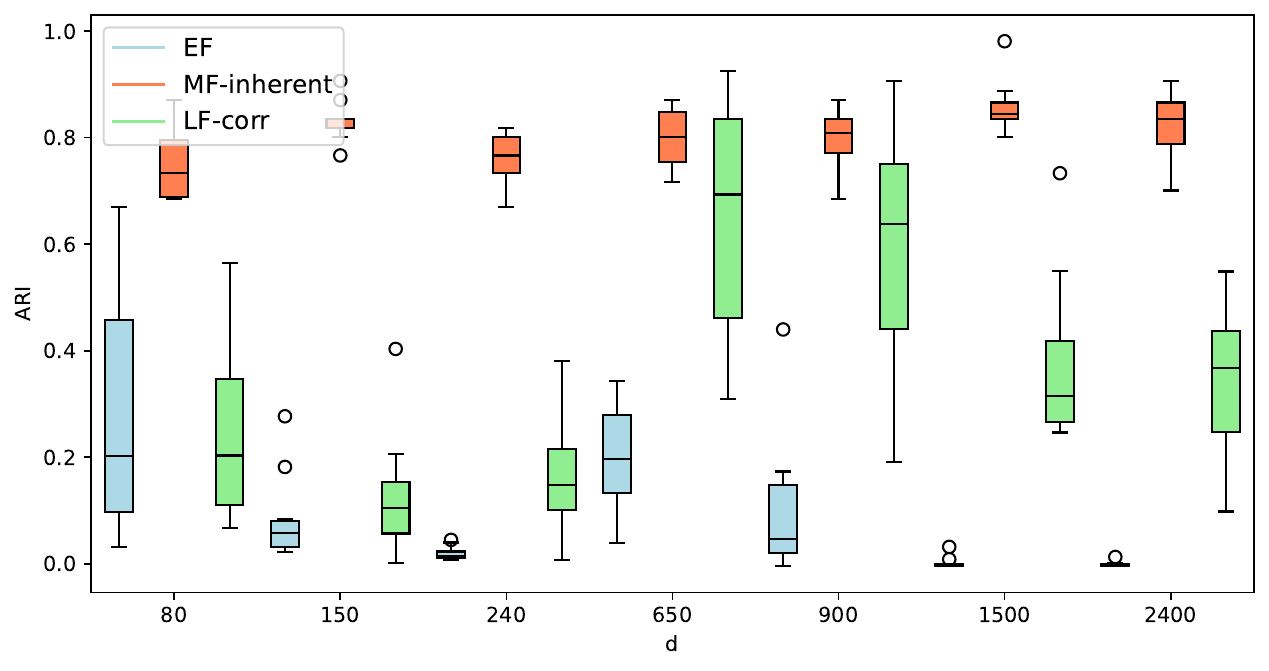}
        \caption{ARI of the clustering models on the data of Example \ref{ex_4}}\label{fig:SC2_boxp}
    \end{subfigure}
		
		\vspace{0.6cm}		
		
    \caption{Experimental results for kernel-based (column 1), NN-based (column 2) and clustering (column 3) models. Row 1 \& 3 show the mean performance over 10 runs, while row 2 \& 4 show the boxplots for the best performing view construction methods per type of fusion. Note that the results for EF in (j) are omitted to improve readability (as the accuracies are much lower).}\label{fig:res}
\end{figure*}

\subsection{Constructed views}

Since Example \ref{ex_2} has an inherent multi-view structure, we can use it to examine the view construction methods. For each value of $d$ we sampled data from Example \ref{ex_2} 10 times and used random split, Euclidean distance-based and Correlation-based feature clustering as feature partitioning methods for view construction. We took the view assignment of all features from the constructed views and calculated the Adjusted Rand Index (ARI) between these and the assignments corresponding to the inherent views. 

\begin{table}
\centering
\caption{ARI between the inherent view assignments of Example \ref{ex_2} and the assignments by the three view construction methods. The mean, std and median are taken over 10 runs and all values of $d$.}
  \begin{tabular}{l|ccc}
   & random & kmeans & corr \\
	\hline 
	mean ARI & $1.7\cdot 10^{-3}$ & $0.93$ & $1.0$ \\
	std ARI &  $\pm 7.1\cdot 10^{-3}$ & $\pm 0.12$ & $\pm 0.0$ \\
	median ARI & $-9.54\cdot 10^{-6}$  & $1.0$ & $1.0$
  \end{tabular}
\label{tab:views}
\end{table}

The results in Table \ref{tab:views} show that Correlation-based feature clustering is always able to identify the inherent view-structure. The Euclidean distance-based feature clustering is able to identify it as well in most cases. As expected, the random split can not approximate the inherent view structure well.

\subsection{Kernel-based classifiers}

The first set of experiments consist of classification using kernel-based methods. For this experiments we used the well known method Least-Squares Support Vector Machines (LS-SVM) \cite{Suykens2002}, and its mid fusion multi-view variant Multi-View LS-SVM (MV-LSSVM) \cite{HOUTHUYS2018b}. For the late fusion variant we considered both average and performance-based fusion where the fusion happens before the $\sign$ function is applied. I.e.:
\begin{equation}
\hat{y}_k(\x_{\mb{t}}) = \sign \left( \sum_v^V \beta_v f_v( \x_{\mb{t}}^{[v]}) \right)
\end{equation}
where $\beta_1, \ldots, \beta_V$ are either $1/V$ (average fusion) or calculated based on the training error covariance matrix (performance-based fusion). The view-specific function $f_v$ represents the prediction before the $\sign$ function is applied, hence:
\begin{equation}
f_v( \x_{\mb{t}}^{[v]}) = \sum_{k=1}^N \alpha^{[v]}_k y_k K^{[v]}(\x_{\mb{t}}^{[v]},\x_{k}^{[v]})+b^{[v]}
\end{equation}
where $\alpha^{[v]}$ and $b^{[v]}$ follow from the dual problem of the model trained on view $v$ and $K^{[v]}$ is the kernel function, which can be chosen differently for each view. 

The hyperparameters tuned are: the regularization parameters (one for early fusion and $V$ view-specific ones for mid and late fusion), the coupling parameter (for mid fusion), the decision function (for late fusion) and the number of views (for mid and late fusion).
To decrease the tuning complexity, we choose the RBF kernel function with kernel parameter $\gamma=1/d$\footnote{Which is the default value in the Python library scikit-learn} for all models and all views.


Figure \ref{fig:kernel_acc} and Figure \ref{fig:kernel_acc_boxp} show the results for Example \ref{ex_1}. It is clear that the accuracy of all methods tends to drop as the dimension increases. MF-corr, however, consistently performs well, even in the extreme case where the dimensionality is more than 10 times the sample size ($N=210$ \& $d=2400$). The boxplots furthermore show the statistical significance of this result. The accuracies of MF-kmeans, MF-random and LF-random are similar to the accuracy of EF, hence using the large feature vector as is, although MF-kmeans outperforms the others by a lot for some values of $d$. LF-corr and LF-kmeans consistently achieve the lowest accuracy. 

Figure \ref{fig:kernel2_acc} and Figure \ref{fig:kernel2_acc_boxp} show the results for Example \ref{ex_2}. In contrast to Example \ref{ex_1}, we can see that all kernel-based models perform relatively well for all values of $d$. Both MF-random and LF-random have similar accuracy as the EF method, while all other multi-view methods outperform these considerably. The boxplots furthermore show that MF-inherent outperforms LF-inherent significantly for higher values of $d$.





\subsection{Neural Network-based classifiers}

\begin{figure}
\centering
\includegraphics[width=0.6\columnwidth]{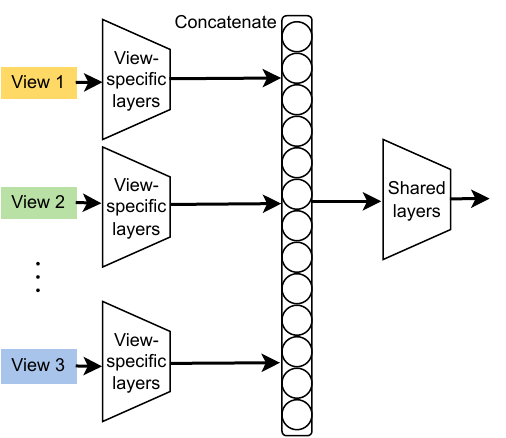}
\vspace{0.1cm}
\caption{Architecture of the mid fusion NN-based classifier}
\vspace{0.4cm}
\label{fig:MF_NN}
\end{figure}

The second set of experiments are carried out using Neural Network (NN)-based classifiers. For this we used standard multi-layer perceptrons with fully connected feedforward layers. The multi-view mid fusion architecture is depicted in Figure \ref{fig:MF_NN}, note that all layers are jointly optimized. All layers except for the last one are followed by a ReLU activation and dropout. The last layer is followed by a sigmoid activation function. For the late fusion variant we considered both average and performance-based fusion where the fusion happens before the threshold is applied.

The hyperparameters tuned are: the number of epochs ($\in \{25,50\}$), the batch size ($\in \{16,32\}$), the dropout rate ($\in \{0,0.5,0.75\}$), the number of layers ($\in \{1,2\}$), the number of neurons in each layer ($\in \{16,32,64,256,512\}$), the number of views (for mid and late fusion) and the decision function (for late fusion). For each model we used the binary crossentropy loss and the Adam optimizer with learning rate $0.001$.


Figure \ref{fig:NN_acc} and Figure \ref{fig:NN_acc_boxp} show the results for Example \ref{ex_1}. It is clear that the accuracy of all methods tends to drop as the dimension increases. The accuracies of all mid fusion methods are similar to the accuracy of EF, none of them significantly outperform one another. The late fusion methods on the other hand, do significantly worse, although the difference in accuracies decreases as the dimension increases.

Figure \ref{fig:NN2_acc} and Figure \ref{fig:NN2_acc_boxp} show the results for Example \ref{ex_2}. While the late fusion and EF methods still show a decrease of performance as the dimension increases, the accuracy of the mid fusion methods stays stable throughout. The boxplots show that these results are significant. The figures furthermore show that the random split performs consistently worse than the other view construction methods.

\subsection{Spectral Clustering}

The last set of experiments involves the unsupervised learning task binary clustering. For this we used the well known method Spectral clustering \cite{Luxburg2007}, and its mid fusion variant Pairwise Co-regularized Multi-view Spectral Clustering \cite{Kumar2011}. Since in an unsupervised setting we do not have access to labels, we did not consider a performance-based late fusion. The decision function for the late fusion version is a two-step process:
\begin{enumerate}
	\item Align the cluster assignment in order to account for permutations. Since we only consider 2 clusters this is easily done by 'flipping' the class assignments and comparing the Manhatten distance between the assignments.
	\item Perform majority voting (average fusion on the cluster assignments).
\end{enumerate}

We used the RBF kernel function with $\gamma=1/d$ to construct the affinity matrices for all models. For the mid fusion method we chose $\lambda=0.02$ in alignment with the results in the work of Kumar et al. \cite{Kumar2011}. For the mid and late fusion we tuned the number of views using the Davies-Bouldin index \cite{DB} as an unsupervised performance metric.


Figure \ref{fig:SC_acc} and Figure \ref{fig:SC_boxp} show the results for Example \ref{ex_3}. It is clear that the mid fusion methods outperform the late fusion and EF methods substantially. Also, while the performance of the mid fusion methods decrease slightly as $d$ increases, the ARI of the late fusion and EF methods drops to almost zero from $d=150$ onwards. There does seem to be a little improvement of using late fusion over EF.

Figure \ref{fig:SC2_acc} and Figure \ref{fig:SC2_boxp} show the results for Example \ref{ex_4}. The figures again show that the mid fusion methods outperform all other significantly. The late fusion models, even though they do not perform as well as the mid fusion ones, do significantly outperform the EF method from $d=150$ onwards.


\subsection{Discussion}

All but one experiment showed that using a multi-view strategy was at least as good as using the large feature vector directly and most showed a significant improvement. The only exception is for the NN-based models on the data of Example \ref{ex_1}, where the late fusion methods underperformed. This might indicate that in this extremely challenging scenario of HDLSS (we know from previous research \cite{Salimi2018} that NN-type models do not do well in an HDLSS setting) and no inherent groupings of features, a more flexible way to fuse information from multiple views is necessary.

Moreover, the mid fusion methods outperformed the late fusion methods in all experiments. Again indicating the need for a more flexible and complex fusion than standard late fusion techniques.

While a simple random split for view construction yield good results for Example \ref{ex_1} \& Example \ref{ex_3} it usually demonstrates weaker performance for Example \ref{ex_2} \& Example \ref{ex_4}. The Euclidean distance- and Correlation-based feature clusterings on the other hand, almost always provided good performance. If we compare this with the results from Table \ref{tab:views}, it indicates that if there is an inherent multi-view structure it is important to try to approximate that with a proper view construction method.



\section{Conclusion and future work}

This paper introduces a universal approach for learning in a high-dimensional low-sample-size (HDLSS) settings using multi-view mid fusion techniques. It shows how existing mid fusion multi-view methods perform well in an HDLSS setting even if no inherent views are provided. Three feature set partitioning methods are demonstrated which split up the high-dimensional feature vectors into multiple views.

Extensive experimental validation across model-types and learning tasks confirm the effectiveness and universality of the approach. Experiments performed with kernel- and NN-based classifiers and spectral clustering methods all showed that mid fusion multi-view learning outperforms the late and early fusion alternatives in an HDLSS setting. Furthermore it was found that if there is an inherent multi-view structure in the data, it is better to use a clustering-based partitioning method instead of a simple random split. 

Given the universal nature of this paper, we believe it can give rise to a wide variety of new methodologies.
Moreover, it lays the groundwork for more research into the universal benefits of using (mid fusion) multi-view learning. Future work can provide a more formal proof of the benefits of mid fusion in an HDLSS setting, and can investigate the mathematical boundaries of this improvement. Furthermore the influence of extra challenges such as extreme class imbalance could be investigated, and the results could be further validated using real-life datasets.



\begin{ack}
The research reported on in this article was partially funded by the Flemish Government under the
Onderzoeksprogramma Artifici\"ele Intelligentie (AI) Vlaanderen programme.
\end{ack}



\bibliography{myBib}

\end{document}